\documentclass[letterpaper]{article}
\usepackage{natbib}
\usepackage{aaailike}
\usepackage{times}
\usepackage{helvet}
\usepackage{courier}
\usepackage[utf8]{inputenc}
\usepackage[T1]{fontenc}
\usepackage{amsmath,amssymb}
\usepackage{amsthm}
\usepackage{graphicx}
\usepackage{booktabs}
\usepackage{multirow}
\usepackage{array}
\usepackage{tabularx}
\usepackage{xcolor}
\usepackage{microtype}
\usepackage{tikz}
\usetikzlibrary{positioning,arrows.meta}
\usepackage{url}
\usepackage[skins,breakable]{tcolorbox}
\theoremstyle{definition}

\definecolor{ourscol}{HTML}{D42300}
\definecolor{bbcol}{HTML}{1868B2}
\definecolor{cbcol}{HTML}{8048AA}
\definecolor{appcol}{HTML}{148843}

\title{Semantic Primes as Explanans for Emotion in Large Language Models}
\author {
    Frank Xing
}
\affiliations {University of Reading\ \ \ \ Email: z.xing@henley.ac.uk}

\begin{document}
\maketitle

\begin{abstract}
Progresses have been made on understanding emotion mechanisms of large language models (LLMs). However, how to explain emotion in LLMs, or even what constitutes good explanations, are less clear. Emotion representations, components, circuits are widely recoverable, but as explanations of a model's own computation they are circular; the emotion space dimensions tend to be arbitrary and non-terminating. A pressing question to ask is whether a more primitive set of internal variables does the work: the semantic primes of the Natural Semantic Metalanguage (NSM). 
Across four instruction-tuned LLMs (Llama-1B, Gemma-2B, Gemma-9B, OLMo-7B), experiments show that the NSM primes are (1) recoverable internal elements; and (2) on the reference model, intervening with a prime based direction controls emotion about three times as strongly, and twice as selectively, as the best appraisal based direction; and (3) the model treats a prime based explication as interchangeable with the corresponding emotion.
These evidences suggest that NSM primes seem to be better explanans for emotion in LLMs than many alternative options according to scientific explanations criteria.
\begin{links}
    \link{Code \& Data}{http://github.com/fxing79/pcs}
\end{links}
\end{abstract}

\begin{figure}[t!]
  \centering
  \includegraphics[width=\linewidth]{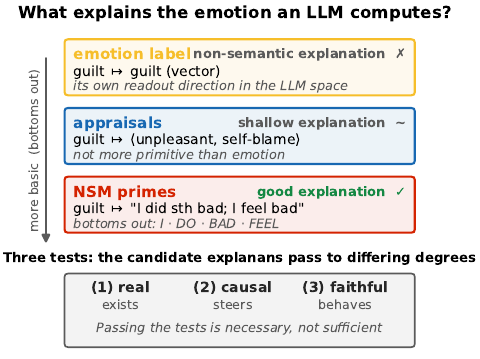}
  \caption{What makes a good causal explanation of the emotion an LLM computes and outoputs? Three tests are drawn from the scientific explanation literature: (1) real: it exists internally, (2) causal: intervening on it moves the emotion, (3) faithful: the model behaves as if it is the emotion. All three candidates pass them, but to differing degrees. What separates them most is the fourth vertical \emph{basicness} axis: an explanans must be more basic than the explanandum. Emotion labels are circular, and appraisals partial reduction; only NSM primes bottom out at a definitional floor.}
  \label{fig:concept}
\end{figure}

\section{Introduction}
Unlike human emotion, which is phenomenal and embodied, emotion in LLMs is functional and computed. The difference calls for a whole set of theory on how emotion emerges and works in LLMs as they are increasingly used and integrated in our society. Progresses have been made on understanding that emotion space~\citep{wu26decod} and emotion representations~\citep{sofroniew26emo} are recoverable in LLMs, and that LLM emotion behavior can be controlled and manipulated via layer injection~\citep{tak2025mechanistic} or circuits~\citep{wang25do}. But there is a significant gap between understanding and (scientific) explanations. For example, I understand ``a magician pulls a rabbit out of a hat'' is a stage effect and is not real, but I cannot explain how he did it. People also have no difficulty understanding an apple falls after ripening without explaining it with plant hormones or gravity. In this sense, there has been little purposeful investigation on explanations of emotion in LLMs from the mechanistic interpretability literature~\citep{bereska2024mechanistic,openmi}. Since using the emotion artifacts (representations, sparse autoencoder extractions, circuits, etc.) to explain themselves is non-semantic and circular, other internal variables must be engaged. Two concept families are often imported from human emotion psychology \citep{tak2025mechanistic}.
The first family of variables are the ``more basic'' discrete emotion states or categories decoded from residual-stream activations as linear directions~\citep{tigges2024linear}, e.g., Ekman's 6 basics or Russell’s Circumplex Model; the second family are continuous ratings, e.g., valence-arousal dimensions~\citep{wu26decod} or Scherer's 21 appraisal dimensions that \citet{tak2025mechanistic} use to probe and steer LLM emotional behavior.

Two problems follow. First, judging the completeness and evaluating competing explanations are intricate. One set of variables may be more linearly recoverable via probing, but that fact alone does not establish better explanations: probe accuracy is correlational and need not reflect causal use \citep{belinkov2022probing}. The mechanistic interpretability field has answered this by adding interventions \citep{vig2020causal,elazar2021amnesic}, and on the narrow technical reading of the word, an explanation is \emph{mechanistic} only when it makes a causal claim \citep{saphra2024mechanistic}. But when another set of variables steers LLM emotion behavior better, weighing between these two sets becomes complicated. 
Second, neither family of variable sets bottom out as explanations. To classify an activation as \emph{guilt} re-applies the annotator's label to the very activation the label
should explain: the explanans is the explanandum. The appraisal space dimensions describe guilt as a profile over a composition of \emph{self\_responsibility} + \emph{unpleasantness}, but those dimensions are not more primitive than the emotions they describe; \emph{unpleasantness} is an affective concept of the same order as \emph{guilt}. The reduction floats and never terminates.

In response to the two problems, I attempt a set of tests to filter out pseudo-explanation and evaluate candidate explanans, and draw a third family of variables from a 60-year-old linguistic program, i.e., the Natural Semantic Metalanguage (NSM) of Wierzbicka and Goddard \citep{wierzbicka1972primitives,wierzbicka1996semantics,goddardwierzbicka2002,goddardwierzbicka2014}. NSM identifies $\sim$65 \emph{semantic primes} (\textsc{want}, \textsc{feel}, \textsc{think},
\textsc{know}, \textsc{do}, \textsc{good}, \textsc{bad}, \textsc{not}, \textsc{because},
\textsc{i}, \textsc{someone}, $\ldots$), held to be mutually indefinable and shared with the
rest of language, so that every other concept, emotions included, can be \emph{explicated}
as a paraphrase built only from primes. For instance, the gold explication of \emph{guilt} is roughly
\emph{``I did something bad; I feel bad because of this; I do not want this to have
happened,''} which bottoms out at primes (\textsc{i}, \textsc{do}, \textsc{bad},
\textsc{feel}, \textsc{want}, \textsc{not}) that are not themselves emotions. This explication of \emph{guilt} is carried as a running example. It is noteworthy that confirming the correctness of NSM in either human language or the LLM case is not this paper's scope. Instead, my claim is an engineering one that NSM primes seem to be better explanans considering the criteria set out here. 

The three families of variables are judged by the standard of causal explanation, on the interventionist account \citep{woodward2003making}: a feature is explanatorily relevant to an outcome when intervening on it changes the outcome. Three demands elaborate (Figure~\ref{fig:concept}), and they come from this account of explanation, not from any tenet of NSM.
\begin{itemize}\itemsep2pt
  \item \textbf{Existence.} The feature is a real internal element, a genuine linear
  representation rather than a probe artifact. This is rigorously tested by decoding above length and
  control-task baselines.
  \item \textbf{Intervention.} Intervening on the feature changes the emotion the model
  computes, and more so the feature serves as an explanan better. This is tested by steering.
  \item \textbf{Behavioral equivalence.} At the input--output level the model treats the
  feature as the concept: in the NSM case, a prime explication is interchangeable with the emotion word and
  is a sufficient cue. This is tested with the model's own inferences and generations.
\end{itemize}
The three demands are necessary: a feature that is absent can be part of the mechanism but is beyond explanations, one that
is causally idle can be a confounder but does not explain, and one that is causally active but means something else
is not the concept~\citep{openmi}. They are not jointly a proof though: a prime's lexical correlate could in principle pass all three, so construct validity remains an open problem
(see Section~\ref{sec:limits}), and the internal mechanism exhibited is a sketch focusing on the raw ingredients rather than a gap-free account.

\paragraph{Contributions.}
Primes inside four instruction-tuned models spanning two orders of magnitude and
three model families (Llama-3.2-1B, Gemma-2-2B, Gemma-2-9B, OLMo-2-7B) are studied to ensure result generalizability. Experiments are built on~\citet{tak2025mechanistic}'s
public harness data so the emotion comparison is head-to-head. 

The result is a single claim that NSM primes are better explanans for emotion in LLMs when compared to other explanans used in literature. It is supported by three findings: (1) primes are real internal elements as 30 of 32 emotion-describing primes are linearly encoded above length and control-task baselines in all four models; (2) on the reference model, a prime direction controls emotion about three times as far, at nearly twice the selectivity, as the best appraisal direction, beating a dimensionality-matched composite ($p<10^{-3}$) and a random control; (3) the model treats a prime explication as interchangeable with the emotion and as a sufficient cue, in all four models and more so than a matched appraisal description. 
For reproducibility, a contrastive suite of 11{,}902 minimal pairs for 32 primes with controls and gold explications is released. Analysis runs on a single CPU and extraction in minutes on a single GPU, so the study is cheap to replicate.

\section{Background}
\subsection{From explanans to their emotion readout}
Mechanistic interpretability seeks a causal, not merely predictive, account of a computation: to explain ``the model reports emotion $e$'' is to exhibit the internal factor that \emph{produces} it. The two requirements discussed in Introduction sort the candidate explanations of emotion by how well each passes the causal tests and how far each reduces. In the emotion label case, a residual activation $h\in\mathbb{R}^d$ at a consolidation layer admits a linear classifier into $K$ possible emotion labels; each emotion is a readout direction $w_e$ and the emotion space is flat, so \emph{guilt} is an atom and the account does not reduce: the explanans is the explanandum. 
In the appraisal case, the same $h$ admits 21 linear regressions onto appraisal scales; \emph{guilt} becomes a profile (2 dimensions activated). 
NSM primes are $\sim$65 directions $v_p$, and an emotion becomes an \emph{explication}, a short structured paraphrase of prime predicates over a floor
shared with the rest of language: \emph{guilt} $\approx$ \emph{``I did something bad; I feel bad because of this.''} The compositional prediction is defined as:
\begin{equation}\label{eq:recipe}
w_e \;\approx\; f\!\Big(\{v_p : p\in P(e)\}\Big),
\end{equation}
with $P(e)$ the prime support of $e$'s explication and $f$ the readout map. The strongest
reading takes $f$ linear, which empirical results reject; the behavioral reading takes $f$ to be whatever the
model computes and that is conceptually the NSM's grammar, which empirical results confirm. Obviously, an explication is structured, not a bag of primes, so a signed sum of prime directions is lossy. Dimensional appraisal is by contrast additive, an emotion a profile over independent dimensions~\citep{smithellsworth1985}; the non-linearity of Scherer's process model \citep{scherer2009cpm} is temporal, not in this static mapping. So the accounts differ in reduction depth and in the form of $f$: on this reading prime composition would be grammatical and appraisal composition dimensional, a contrast the steering experiment puts to the test.

\subsection{Primes and emotion in LLMs}
That LLMs present and process semantic primes seems self-evident, though the proof is absent from literature, especially whether those are used when processing emotion. A recent study confirms prime-related schema slots from LLM outputs~\citep{faithbydef2026}. At mid-layers, emotion is read from residual streams as linear structure~\citep{tak2025mechanistic} or extracted using a sparse autoencoder~\citep{wu26decod}. The work closest to this one is \citet{tak2025mechanistic}, who probe 13 emotion categories and 21 appraisal dimensions across 10 open LLMs, localize mid-layer emotion heads by activation patching and knockout, and steer behavior with orthogonalized appraisal injection; this pipeline is reproduced as the comparability anchor. Since primes are fundamental concepts, existence test adds a Hewitt-Liang control task \citep{hewittliang2019probes} to separate a represented feature from probe capacity, within the linear-representation framing of \citet{park2024linear}; sparse autoencoders \citep{lieberum2024gemmascope,templeton2024scaling} are an alternative route used only in pre-trained form; and an LLM pipeline that generates and verifies NSM explications \citep{deepnsm2025} supplies those gold recipes not published in literature.

\section{A Contrastive Suite for Semantic Primes}
\label{sec:suite}
Probing primes needs stimuli that assert a single prime while holding lexical context fixed. For this reason, a contrastive minimal-pair suite for 32 primes is built and released with gold explications, code, and a machine-translated Chinese version for cross-lingual studies. 
For each prime templated pairs were generated: a positive member asserting the prime and a negative member with the same content but the prime absent. For example for \textsc{want}, 

``\emph{(+)The nurse wants to play the piano}''

``\emph{(-)The nurse played the piano}''; 

\noindent and for \textsc{not}, 

``\emph{(+)The driver did not open the gate}'' 

``\emph{(-)The driver opened the gate}.'' 

\noindent Templates are paraphrased under rejection rules that forbid the prime's exponent word in the negative member, match target word-length within $\pm1$ where feasible, and track a syntactic-template identifier for a held-out split. The 32 primes span the symbol classes that admit a single-prime contrast: mental predicates (\textsc{want, feel, think, know, say, do}), evaluators (\textsc{good, bad}), modals (\textsc{can, maybe, not, true}), person primes (\textsc{i, someone, other, people}), temporal, descriptor, bodily, and reason primes.

The selection of primes is principled, not a convenience sample. A contrastive probe needs a prime a sentence can assert or withhold while holding context fixed, which is natural for predicates and operators but ill-posed for referential substantives and determiners (\textsc{this}, \textsc{something}, \textsc{kind}) that appear in nearly every sentence, and for quantifiers (\textsc{one}, \textsc{some}, \textsc{all}) whose contrast is graded and noun-phrase-bound.
Crucially, of the 22 distinct primes that appear in the gold explications of the 13 target emotions, the suite covers 21; the lone exception is \textsc{this}. The untested primes are dominated by the quantifier and spatial classes, which appear in no emotion explication, so the 32 primes already cover every emotion experiments required in this paper. The full prime inventory is in Appendix~\ref{app:suite}.

The suite contains 11{,}902 minimal pairs / 23{,}804 sentences, $\ge$202 pairs per prime, and
$\ge$16 held-out template pairs per prime. No negative leaks the prime's exponent (a hard
generation constraint); the length confound is quantified per prime as Cohen's $d$ on word
count, and the probing harness always reports a length-only baseline, so a prime counts as
encoded only when it clears the length shortcut. The pipeline is validated on synthetic data,
successfully recovering a planted layer for six synthetic primes with off-layer accuracy at chance (Appendix~\ref{app:suite}).

\section{Experimental Setup}
\label{sec:setup}
\paragraph{Models and data.}
Four LLMs are studied: Llama-3.2-1B-Instruct (16 blocks, $d{=}2048$),
Gemma-2-2B-it (26 blocks, $d{=}2304$), Gemma-2-9B-it (42 blocks, $d{=}3584$), and
OLMo-2-7B-Instruct (32 blocks, $d{=}4096$), spanning two orders of magnitude and three
families, including the fully open OLMo whose independent pre-training makes a cross-family
claim more than a within-lineage one. Llama-3.2-1B is the reference model for the intervention experiment because of its quasi-linear emotion readout; the other three test prime existence and explication behavioral equivalence across family and scale. Emotion and appraisal labels come from crowd-enVent \citep{troiano2023crowdenvent} (6{,}800 event descriptions annotated for 13 emotions and 21 appraisal dimensions); I reproduce \citet{tak2025mechanistic}'s split of 2{,}740 train and 1{,}370 test sentences, with ISEAR \citep{scherer1994isear} as an additional robustness check.

\paragraph{Probes and steering.}
Linear probes are L2-regularized logistic regression, appraisal probes L2 ridge; hyperparameters are not retuned per layer. Steering adds a unit-normalized direction, scaled by the layer's mean residual norm, to the block residual at layer~11, the mid-network consolidation site fixed by the patching peak (Section~\ref{sec:reduce}), at all token positions. The behavioral tests use only the model's generations and answer-token logits, with no probe, and run on all four models. For existence, this study reports per-prime peak accuracy against the length baseline and Hewitt-Liang selectivity; for intervention, target-logit shift, off-target mass, selectivity, and target-argmax rate; for behavioral equivalence, explication-to-word logit agreement, explication-only classification with prime ablation, and the linear recipe match
. Cached activations are reused and never recomputed.

\section{Primes Are Real Internal Elements}
\label{sec:exist}
This section presents existence test results: a prime must be a genuine internal element, not a probe
artifact. Decoding establishes this only with controls. Probe accuracy alone only shows a label is
recoverable, while clearing a Hewitt-Liang control task shows the model represents the prime~\citep{hewittliang2019probes}, since the
control measures what a probe of the same capacity can fit on structureless pseudo-labels.

\paragraph{Protocol.}
For each of the 32 primes, residual activations are extracted on its contrastive pairs (balanced,
last token, every layer) and fit one logistic probe per layer, recording a length-only baseline,
a Hewitt-Liang control-task probe \citep{hewittliang2019probes}, and a held-out template split.
A prime counts as \emph{linearly encoded} when its peak accuracy clears the length baseline significantly and shows selectivity over other primes.

\begin{figure}[b]
  \centering
  \includegraphics[width=0.88\linewidth]{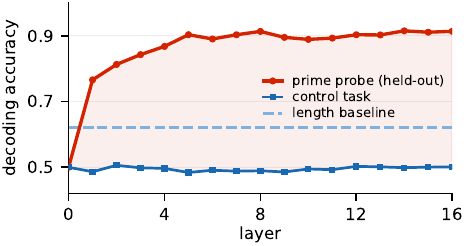}
  \caption{Existence on Llama-3.2-1B: per-layer prime decoding, averaged over the 32 primes. Held-out
  template accuracy (red) rises to $0.91$ and stays far above the Hewitt-Liang control-task accuracy
  (grey, near the chance line $0.5$) and the length-only baseline (dashed); the shaded gap is the
  selectivity that separates a represented prime from a probe artifact.}
  \label{fig:decod}
\end{figure}

\paragraph{Existence and replication.}
Thirty of the 32 primes are linearly encoded in all four models: 31/32 on Llama-3.2-1B (bootstrap 95\%
CI $[29,32]$), 30/32 on Gemma-2-2B, and 32/32 on Gemma-2-9B and OLMo-2-7B, across three families and the
1B-to-9B scale range. Averaged over the 32 primes on Llama-3.2-1B, held-out template decoding reaches
$0.91$, far above the Hewitt-Liang control task ($0.50$, at chance) and the length-only baseline
($0.62$); the gap between probe and control is the selectivity that makes this an existence claim and not
a probe artifact (Figure~\ref{fig:decod}). The two exceptions across models, \textsc{do} and \textsc{say},
carry the worst length confound ($d{=}1.26$ and $2.03$); other confounded primes (\textsc{know},
\textsc{can}, \textsc{think}, all $d>1.2$) still clear the bar, so the control separates genuine encoding
from a length shortcut. The atoms \emph{guilt}'s explication is built from, \textsc{i}, \textsc{do},
\textsc{bad}, and \textsc{feel}, are all among the encoded primes, so the ingredients its recipe needs are
all present in LLMs. Decoding with controls, replicated across four models, establishes existence but not use: a probe
can read a direction the model never uses, which the next section tests by intervention.

\section{Primes Control Emotion Better}
\label{sec:steer}
This section presents intervention test results. In short, a direction assembled from prime atoms and injected into the residual stream is found to control emotion more strongly and more selectively than the best appraisal based direction. The shift is about three times as large at nearly twice the selectivity, the gap widest been on the agency axis (i.e.,  \textsc{i} versus \textsc{someone}), suggesting a less effective modeling construct by appraisals.

\paragraph{One mechanism, six directions.}
A forward hook adds a vector to the block residual at layer~11, at all token positions. The six arms differ only in the injected direction: the \emph{emotion} probe direction $w_e$ (a readout ceiling); the \emph{single appraisal} most associated with the target (e.g., guilt to
\emph{self\_responsblt}, anger to \emph{other\_responsblt}, joy to \emph{pleasantness},
sadness to \emph{unpleasantness}); a \emph{multi-appraisal composite}, the top-$k$ appraisals with
$k$ matched to the recipe's component count; the \emph{single best prime}; the \emph{prime recipe},
a centrality-weighted signed sum of fitted prime directions (guilt as
$\textsc{i}+\textsc{do}+\textsc{bad}$, so is still lossy); and a \emph{random-concept} control. Every direction is
unit-normalized and scaled by the layer's mean residual norm, so the dose is a common fraction of
residual magnitude; the composite and random arms control for direction count and for injecting any
vector at all. The 13 emotion-token logits are read at the answer position over four targets (guilt,
anger, joy, sadness, see Table~\ref{tab:steer-pertarget}).

\begin{table}[b]
  \centering\small
  \begin{tabular}{l rr rr rr}
  \toprule
  & \multicolumn{2}{c}{emotion} & \multicolumn{2}{c}{appraisal} & \multicolumn{2}{c}{prime} \\
  Target & shift & sel & shift & sel & shift & sel \\
  \midrule
  guilt   & $17.34$ & $0.83$ & $0.41$ & $0.13$ & $3.75$ & $0.55$ \\
  anger   & $12.73$ & $0.78$ & $1.82$ & $0.38$ & $6.60$ & $0.66$ \\
  joy     & $14.29$ & $0.77$ & $4.69$ & $0.59$ & $4.67$ & $0.62$ \\
  sadness & $10.40$ & $0.72$ & $1.25$ & $0.33$ & $4.61$ & $0.56$ \\
  \bottomrule
  \end{tabular}
  \caption{Per-target steering at $\beta=2$ (Llama-3.2-1B). The emotion arm is the readout ceiling; the prime
  arm beats the appraisal arm on every target emotion.}
  \label{tab:steer-pertarget}
\end{table}

\begin{figure}[t]
  \centering
  \includegraphics[width=\linewidth]{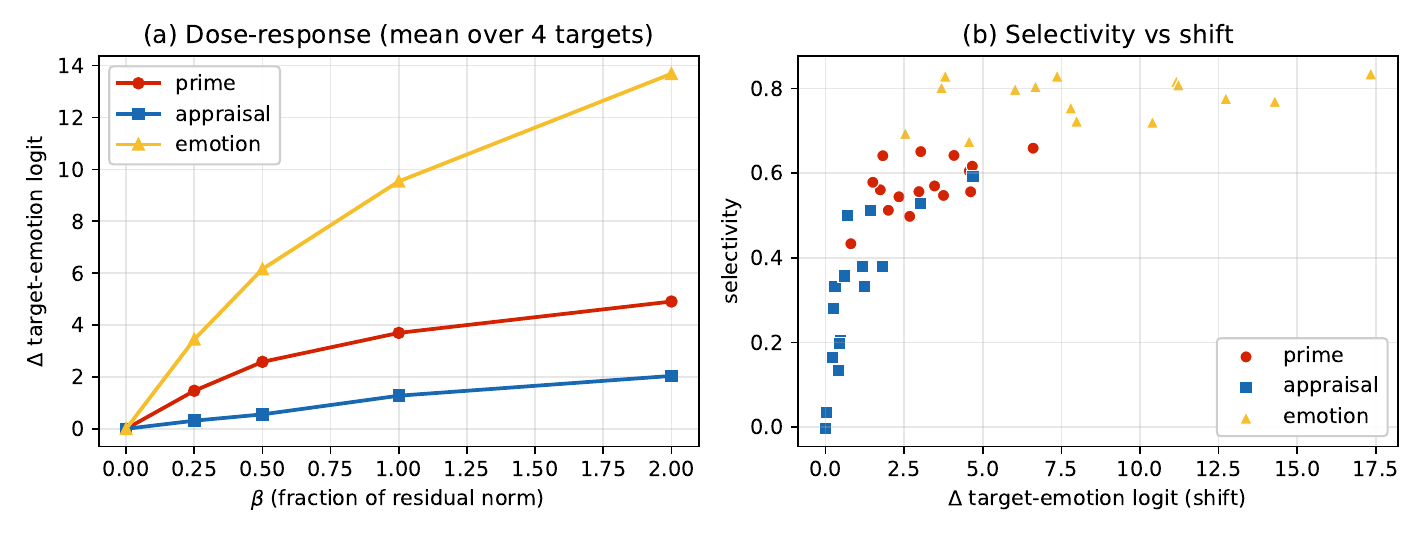}
  \caption{Steering head to head (Llama-3.2-1B): prime (red), appraisal (blue), emotion (yellow).
  (a) Mean target-logit shift versus dose. (b) Selectivity versus shift for all positive doses; upper
  right is better. The prime arm dominates the appraisal arm on both axes; the emotion arm is the readout ceiling -- best control but weakest explanatory power.}
  \label{fig:steering}
\end{figure}

\paragraph{Per-target steering detail.}
Averaged over the four targets and the positive dose grid ($\beta\in\{0.5,1,2\}$), the prime recipe
shifts the target emotion by $3.73$ logits at selectivity $0.58$, while the single best appraisal
shifts it by $1.29$ at $0.31$ (Table~\ref{tab:steer}, Figure~\ref{fig:steering}). Against the
component-matched composite the prime recipe still shifts emotion further ($3.73$ versus $2.11$
grid-averaged; strongest dose $4.91$, CI $[4.76,5.05]$, versus $2.84$, $[2.68,2.99]$) and more
selectively ($0.60$ versus $0.54$, non-overlapping); a permutation test gives a $+2.07$-logit gap,
$p<10^{-3}$. The random-concept control barely moves the target ($0.02$, selectivity $-0.26$), a
near-zero causal floor, and a single best prime already about matches the full recipe ($3.62$ at
$0.61$), so the handle is carried by the prime representation at matched dimensionality, not by
component count. The advantage is sharpest on the agency axis (selectivity $0.57$ for primes versus
$0.27$ for the single appraisal); for anger the prime arm drives anger to the top prediction in
$77\%$ of prompts. Injecting the emotion direction itself shifts the target most ($9.80$, selectivity
$0.78$), as expected for the readout direction; the prime arm, built from general-purpose atoms,
recovers a large and selective fraction of that ceiling.

\begin{table}[t]
  \centering\small
  \begin{tabular}{l r r}
  \toprule
  \multicolumn{3}{@{}l}{\emph{Llama-3.2-1B}, over four targets $\times$ dose grid} \\
  Arm & shift & selectivity \\
  \midrule
  emotion (ceiling)            & $9.80$ & $0.78$ \\
  \textbf{prime recipe (ours)} & $\mathbf{3.73}$ & $\mathbf{0.58}$ \\
  prime, single best           & $3.62$ & $0.61$ \\
  appraisal, multi (matched)   & $2.11$ & $0.41$ \\
  appraisal, single            & $1.29$ & $0.31$ \\
  random (control)             & $0.02$ & $-0.26$ \\
  agency axis, prime     & $...$ & $0.57$ \\
  agency axis, appraisal    & $...$ & $0.27$ \\
  \midrule
  \multicolumn{3}{@{}l}{\emph{Gemma-2-9B}, grid-average shift} \\
  Composition & linear & non-linear \\
  \midrule
  prime (NSM) & $0.02$ & $\mathbf{0.97}$ \\
  appraisal   & $0.74$ & $0.41$ \\
  \bottomrule
  \end{tabular}
  \caption{Steering, two models. Llama: the prime recipe beats both appraisal arms on shift and
  selectivity (gap versus composite $+2.07$, $p<10^{-3}$); random control near zero; emotion ceiling
  $9.80$. Gemma-9B: the model's own explication encoding steers primes (mostly via content) but not
  appraisals; ceiling $1.82$, floor $0.10$.}
  \label{tab:steer}
\end{table}

\paragraph{Robustness check on injection layer.}
The prime advantage is not an artifact of a particular direction fit or injection site. It holds in all
six bootstrap resamples of the probe-fitting data, where the appraisal arm varies but the prime arm stays
ahead, and at every injection layer from~7 to~14, where all directions including the prime recipe are
refit, the prime arm leads. The steered direction is also nearly orthogonal to the emotion readout: the cosine between
the prime recipe and the emotion probe direction $w_e$ averages $0.04$ across targets. A direction
geometrically unrelated to the readout still drives the emotion logit, which is the signature of a
\emph{compositional} causal effect: the primes are injected upstream and processed by the network into
the emotion, not read off the recipe directly. This also reconciles with the geometric non-reduction of
Section~\ref{sec:reduce}, where the emotion readout is not a linear function of prime directions. The
composition is robust to the ingredient: assembled from the model's own single-prime encodings, a
contrastive difference of means, rather than from probe directions, the recipe steers as well ($4.5$
versus $3.9$ at matched selectivity $0.59$), and both far exceed injecting the whole explication's
encoding ($1.1$, selectivity $-0.25$). Independent prime vectors, composed, out-steer the holistic
sentence encoding, so the effect is neither a probe-geometry artifact nor a global-content injection.

\paragraph{On larger models the linear recipe does not transfer.}
The linear prime recipe is model-specific. With per-model dose and layer calibration it is inert on
Gemma-2-9B (grid-averaged shift $0.02$, at the random floor), and assembling it from the model's own
single-prime encodings rather than probe directions does not rescue it ($0.14$): the primes do not
linearly compose there by either ingredient. The emotion is reachable only from the whole explication's
encoding ($0.97$ at selectivity $0.66$, against an emotion-readout ceiling of $1.82$), the trivial content
direction any model carries for guilt-describing text. Phrased in primes that content still out-steers a
matched appraisal description ($1.58$ versus $0.74$), and the same treatment leaves appraisal at its
additive profile ($0.74$ versus $0.41$; Table~\ref{tab:steer}). 
For this reason, the main analysis uses Llama-3.2-1B, where the compositional recipe itself steers; on Gemma-2-9B the prime explication out-steers the appraisal description as content, but the linear prime composition does not transfer. On OLMo-2-7B the readout ceiling is ill-set (a negative emotion-arm shift), so numbers are uninformative and not reported.

\section{Emotions Reduce to Primes Behaviorally}
\label{sec:reduce}
This section presents behavioral equivalence test results. In all four models the model treats a prime explication as
the emotion. The reduction is behavioral, not geometric: the late readout is not a linear sum of prime
directions, and the localization of emotion within the network explains why the two levels diverge.

\paragraph{Behavioral equivalence and sufficiency.}
With no probing and on all four models, the 13 emotion-token logits are read for: (1) the emotion word, (2) its gold explication, (3) a same-affect decoy explication, e.g.,~shame for guilt, differing by one \textsc{people-know} component, and (4) a prime-scrambled control. Given the explication alone, with no emotion word present, the model classifies it as the target emotion well above the 13-way chance rate of $0.08$, at $0.54$, $0.85$, $0.77$, and $0.54$ respectively for Llama-3.2-1B, Gemma-2-2B, Gemma-2-9B, and OLMo-2-7B, and the target explication beats its same-affect decoy by $0.30$ to $0.63$ probability mass, so it is not riding on generic affect words or on the presence of an emotion label, which the explication never contains. Reduction is also sufficient: ablating the explication one prime at a time, removing a \emph{central} prime degrades the target logit more than removing a \emph{peripheral} one in every model ($0.59$ versus $0.12$ on Llama-3.2-1B, and the same ordering elsewhere; see Table~\ref{tab:behav}).

\begin{table}[ht]
  \centering\small
  \setlength{\tabcolsep}{4pt}
  \begin{tabular}{l r r r r}
  \toprule
  & Llama & Gem.2B & Gem.9B & OLMo \\
  \midrule
  \multicolumn{5}{@{}l}{\emph{prime explication}}\\
  (E) expl$\to$emotion acc   & $0.54$ & $0.85$ & $0.77$ & $0.54$ \\
  (E) decoy margin           & $0.30$ & $0.60$ & $0.63$ & $0.35$ \\
  (S) central drop           & $0.59$ & $0.14$ & $0.18$ & $0.51$ \\
  (S) peripheral drop        & $0.12$ & $-0.21$ & $0.03$ & $-0.14$ \\
  \midrule
  \multicolumn{5}{@{}l}{\emph{appraisal description}}\\
  (E) expl$\to$emotion acc   & $0.25$ & $0.33$ & $0.33$ & $0.42$ \\
  (E) decoy margin           & $0.03$ & $0.10$ & $0.03$ & $0.11$ \\
 (S) central drop           & $0.68$ & $0.92$ & $-0.24$ & $-0.27$ \\
 (S) peripheral drop        & $0.35$ & $0.32$ & $0.03$ & $0.31$ \\
  \midrule
  floor: prime              & $0.47$ & $0.21$ & $0.34$ & $0.40$ \\
  floor: non-prime          & $0.73$ & $0.50$ & $0.72$ & $0.72$ \\
  \bottomrule
  \end{tabular}
  \caption{Behavioral test batteries, all four models. 13-way chance for classification is $0.08$. The explications are read as the emotion (E) and are compositionally sufficient (S, central $>$ peripheral in every model) more strongly than a matched appraisal description, whose decoy margin is near zero and
  whose sufficiency gradient inverts on the two larger models. Primes resist simplification more than
  non-primes in every model (floor).}
  \label{tab:behav}
\end{table} 

For \emph{guilt}, the evaluative \textsc{bad} and \textsc{feel} are central and the \textsc{want}/\textsc{not} clause peripheral: the central primes carry the emotion. The same battery on a matched appraisal description, a Component-Process-style clause profile built from the crowd-enVent appraisal ratings, is markedly weaker (Table~\ref{tab:behav}): the description is read as the
target emotion below the prime explication on every model ($0.25$ to $0.42$ versus $0.54$ to $0.85$),
beats its same-affect decoy by only $0.03$ to $0.11$ against the prime's $0.30$ to $0.63$, and its
central-clause sufficiency inverts on the two larger models. The model treats a prime explication, not a
matched appraisal profile, as interchangeable with the emotion, on our best operationalization of each.
A fluency confound does not explain this gap. On the reference model the prime explication is more fluent
than the matched appraisal description (mean per-token log-probability higher by $0.81$ nats, $p<10^{-3}$),
yet under fluency-invariant scoring, domain-conditional PMI \citep{holtzman2021surface} and contextual
calibration \citep{zhao2021calibrate}, the prime advantage persists and widens, because calibration removes
a label-prior bias that suppressed the explication.

\paragraph{The readout is not a linear sum.}
The prime directions are fit from the contrastive suite in the same residual space as the emotion vectors
and build a centrality-weighted recipe direction per emotion from gold explications
\citep{wierzbicka1999emotions}; the match test asks whether $w_e$ is closer to its own recipe than to
the other twelve emotions. Both direct sources land near chance (Table~\ref{tab:decomp}): out-of-context rank-3
$0.39$ against chance $0.23$, in-context exactly at chance. Emotion is linearly recoverable from the
prime coordinates, but not because of them: projecting the
consolidation-layer residual onto the 32 prime directions and training a classifier reaches $0.86$,
yet a \emph{random} 32-dimensional projection reaches $0.82$, so the prime subspace adds only $0.05$,
and per-prime ablation moves accuracy by at most $0.02$. The prime subspace is not a privileged basis.

\paragraph{A mechanism sketch.}
The two findings reconcile through localization. The emotion computation peaks at layer~10, and by
the consolidation layer the prime content has been \emph{consumed} into the emotion representation: of
18 primes tested at the answer position, only \textsc{feel} (lift $+0.23$) and \textsc{someone}
($+0.11$) remain decodable. The late readout cannot be a linear sum of prime directions because there
are almost no prime directions left to sum, and the survivors are in a sense, the emotion-handles: the
feeling predicate and the other-person agency role. Emotions reduce to primes as a computation the
model carries out in the middle layers and spends, not as a static linear geometry of the final state.
I present this as a mechanism sketch: it localizes where the reduction happens but does not trace the
layer-10-to-readout step arrow by arrow.

\section{Do Primes Bottom Out in LLMs?}
\begin{figure}[t]
  \centering
  \includegraphics[width=0.95\linewidth]{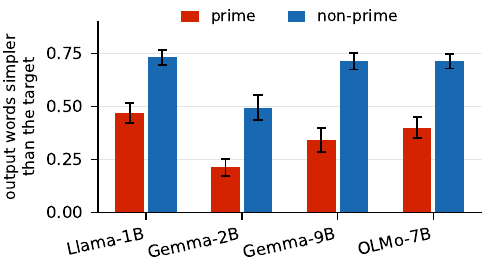}
  \caption{The definitional floor, behaviorally. Asked to restate a word in simpler terms, the model
  produces a smaller fraction of simpler (shorter, higher-frequency) words for a prime than for a length-
  and class-matched non-prime, in all four models: a prime resists reduction, a non-prime does not. Error
  bars are the standard error of the mean over the primes and the matched non-primes.}
  \label{fig:floor}
\end{figure}

The explication reduces an emotion to primes; a probe-free signature shows the primes themselves do not
reduce further. Prompted to restate a word ``using only simpler words,'' the model simplifies a non-prime
far more readily than a prime: the fraction of output words more frequent than the target is
$0.73/0.50/0.72/0.72$ for non-primes against $0.47/0.21/0.34/0.40$ for primes (see Table~\ref{tab:behav} and Figure~\ref{fig:floor}). A prime resists reduction in a sense that more than half of the explaining words cannot be simpler, so the explication regress terminates. This is the bottom-out property an explanation demands: \emph{guilt} reduces to \textsc{i}, \textsc{do}, \textsc{bad}, and \textsc{feel}, which are not to be explained further in the semantic space, where a label or an appraisal dimension would still float.

The localization of primes also seems earlier than emotions. I reproduce and concur the localization of emotions on Llama-3.2-1B by \citet{tak2025mechanistic}, which fixes the consolidation band and the injection layer. Emotion accuracy saturates late (linear probe peak $0.943$ at layer~13); activation patching is sharper, transplanting the event-position residual flips the target prediction at a rate that climbs through the mid-network and peaks at $0.805$ at layer~10 before falling, recovering a mid-layer attention-mediated consolidation. It is this peak that fixes layer~11 as the injection site. Full curves and the emotion geometry, where valence dominates and the agency
contrast is secondary, are in Appendix~\ref{app:loc}.

\section{Discussion}
\label{sec:discussion}
This paper shows the potential of semantic primes as a handle that explains in LLMs, not predicts over complicated theories. The two standard accounts of emotion in LLMs do not bottom out: labels relabel the activation, appraisals describe it in terms no more primitive than emotion, and both are validated mostly by probes that show recoverability, not use. Judged by what it takes to explain the emotion a model computes, primes pass where these fall short: they are real internal elements, the model treats their explications as the emotion, and intervening on them controls emotion more than appraisals do. The same primes (\textsc{want}, \textsc{not}, \textsc{can}, \textsc{feel}) appear in explications of decisions, permissions, and requests, so the prime-interface is more universal than emotion-specific explanans.

It worths mentioning, however, that geometrically the emotion direction is not a linear sum of prime directions; though behaviorally the model treats an emotion word and its explication as interchangeable and the explication as sufficient. A causal handle need not be a geometric basis. The reconciliation is mechanistic: primes are computed in the mid layers (e.g., 7/16) and consumed into the emotion representation (11/16), so the final state retains only the feeling and agency atoms. A
prime-indexed account of emotion is therefore viable, but it must read emotion off the primes nonlinearly.

The clean compositional result is the Llama one, and it is strong there: independent prime vectors, whether probe directions or the model's own single-prime encodings, compose into a direction that out-steers the whole explication's encoding, so the effect is composition from parts, not a global-content injection. On Gemma-2-9B the parts do not compose by either ingredient; only the whole-content direction steers, which is the trivial encoding any model has for emotion-describing text. The prime explication still out-steers a matched appraisal description there, and the appraisal arm gains nothing from the same treatment, so the content asymmetry is real cross-model even though the compositional claim is Llama-only. Tracing whether a larger model composes primes through a different circuit is left open (Section~\ref{sec:limits}).

One structure recurs as a causal signal even though valence dominates the static geometry. Agency is
where prime steering most beats appraisal (selectivity $0.57$ versus $0.27$), the one decomposition that
survives across direction sources (anger at rank-1), and \textsc{someone} is one of only two primes still
decodable at the readout. In NSM it is a single substitution, $\textsc{i}\leftrightarrow\textsc{someone}$,
inside a recipe of the form \emph{X did something bad; I feel bad because of this}, the minimal edit that
turns guilt into anger.

\section{Limitations and Conclusion}
\label{sec:limits}
The main limitation is on construct validity. Because the three tests are necessary for a set of variables to count as an explanation, but not complete or jointly a proof, it is possible that a decoded direction measures a lexical correlate, not the NSM prime per se. More external validation can be supplemented by blind human paraphrases written by annotators unaware of the target prime.

A second limitation is on the full understanding from prime ingredients to emotion representations: the mechanism is only a sketch. The intervention is clean on Llama-3.2-1B and transfers to Gemma-2-9B under non-linear composition (Section~\ref{sec:steer}); but OLMo-2-7B is not calibrated, and the non-linear appraisal arm uses a constructed appraisal description that a stronger operationalization could potentially improve, so the cross-model causal claim rests on two models. The mechanism that reconciles behavioral and geometric reduction localizes a mid-network consolidation band but does not trace the layer-10-to-readout step. Concretely, the attention heads or MLP sub-modules that compose the primes, or the NSM grammar component, are unidentified.
The gold recipes are taken from the published NSM canon \citep{wierzbicka1999emotions,goddardwierzbicka2014}, whose explications carry the validity of a long peer-reviewed program; what is specific to us is the reduction to a 32-prime working subset and the centrality weighting used for ablation. Behavioral robustness to that operationalization can be further checked by recipe-perturbation or re-deriving the central-versus-peripheral ablation under alternative but NSM-valid explications.

Another limitation is that cross-lingual universality of primes is not tested in the LLM case, though it is one of the NSM proposition for human language. Transfer to other languages would be confounded by a multilingual model routing other languages through an English-like latent space \citep{wendler2024llamas,schut2025multilingual}, so separating universality from English-pivot routing requires native-authored stimuli in a typologically distant language and ideally an individually trained non-English LLM, which is left to future work.

In summary, this research confirms semantic primes to be good explanans of emotion in LLMs and requires minimal theory-building. Although other models~\citep{tak2025mechanistic}, linguistic features~\citep{yang25emo} or narrative style~\citep{shen24heart} may as well explain emotion, they are likely correlational (not causal) and subject to further specification. It also presents a framework for assessing the quality of scientific explanations, therefore makes methodological contribution to mechanistic interpretability.

\bibliography{bib/aaai.bib}

\clearpage
\appendix
\section{The Contrastive Suite: Inventory, Confounds, Synthetic Validation}
\label{app:suite}
\paragraph{Symbol classes and the full inventory.}
The 32 tested primes are mental predicates (\textsc{want, feel, think, know, say, do}), evaluators
(\textsc{good, bad}), modals (\textsc{can, maybe, not, true}), person primes (\textsc{i, someone, other,
people}), temporal (\textsc{now, before, after, a-long-time, a-short-time}), descriptors (\textsc{big,
small, very, more, same}), bodily and animate (\textsc{body, live, die}), and reason (\textsc{because, if,
happen}). The untested $\sim$33 are dominated by quantifiers (\textsc{one, two, some, all, much, few}),
spatial primes (\textsc{here, above, below, near, far, inside, touch, side, where}), and the referential
substantives and determiners (\textsc{you, something, this, kind, part}); none of these appears in a gold
emotion explication, and none admits a clean single-prime assertion-contrast. Of the 22 distinct primes used
across the 13 emotion recipes, 21 are tested (only \textsc{this} is missing).

\begin{figure}[h]
  \centering
  \includegraphics[width=\linewidth]{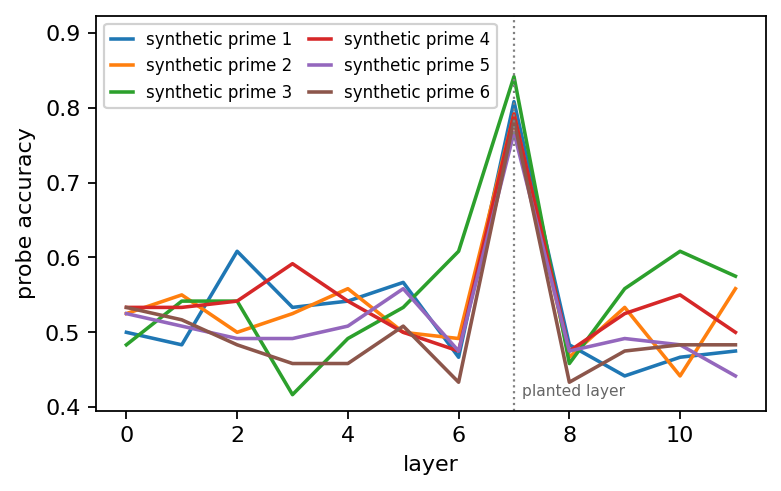}
  \caption{Synthetic validation: a prime-style signal planted at layer~7 of a random-initialized 12-layer
  transformer is recovered for all six synthetic primes (the spike); off-layer accuracy and length baseline at chance.}
  \label{fig:synthetic}
\end{figure}

\paragraph{Splits, confounds, validation.}
Split is done 70/30 at the template level. Negated sub-sets (\textsc{want, know, can, true}) and graded sub-sets
(\textsc{time3, valence3, size3, duration2}) test partial orderings inside a prime family. Zero negatives leak
the prime's exponent; 13 primes show $|d|>0.5$ on word count, the most extreme \textsc{say} at $d{=}2.03$. The
harness recovers a planted layer for all six synthetic primes with off-layer accuracy and length baseline at
chance (Figure~\ref{fig:synthetic}).

\clearpage
\section{Localization: Full Curves and Geometry}
\label{app:loc}
\begin{figure}[h]
  \centering
  \includegraphics[width=\linewidth]{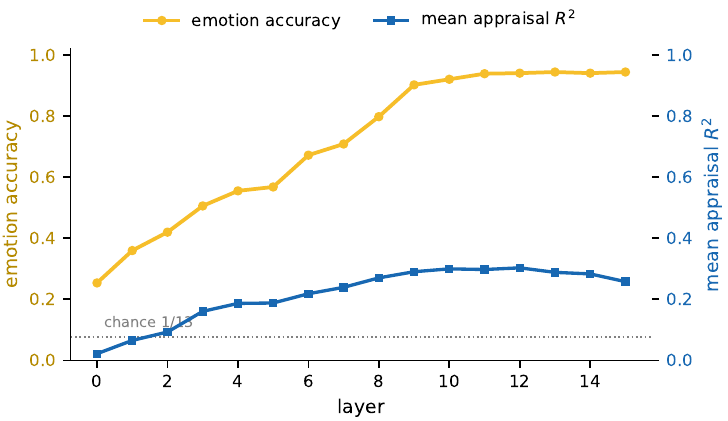}
  \caption{Layer-wise probing of Llama-3.2-1B on crowd-enVent. Emotion accuracy (left) and mean appraisal $R^2$
  (right) saturate late, fixing the consolidation band; chance accuracy is $1/13 = 0.08$.}
  \label{fig:layersweep}
\end{figure}

\begin{figure}[h]
  \centering
  \includegraphics[width=\linewidth]{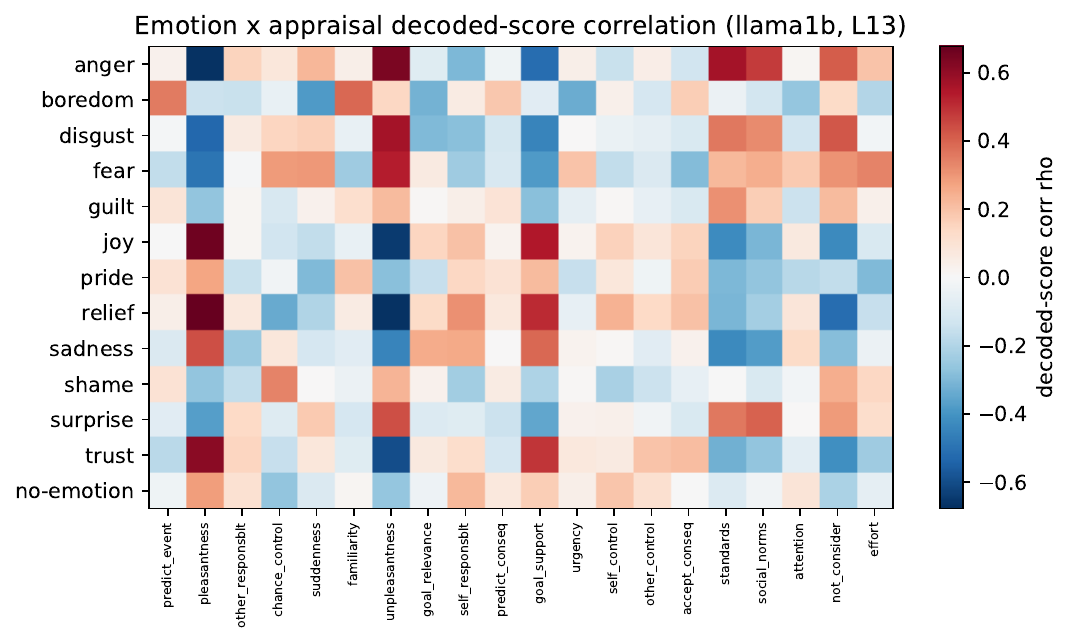}
  \caption{Decoded-score correlation $\rho$ between the 13 emotion readouts and the appraisal readouts at
  layer~13. Valence dominates; the agency contrast is fainter, visible as anger loading on other- over
  self-responsibility.}
  \label{fig:emoapp}
\end{figure}

\begin{figure}[t]
  \centering
  \includegraphics[width=\linewidth]{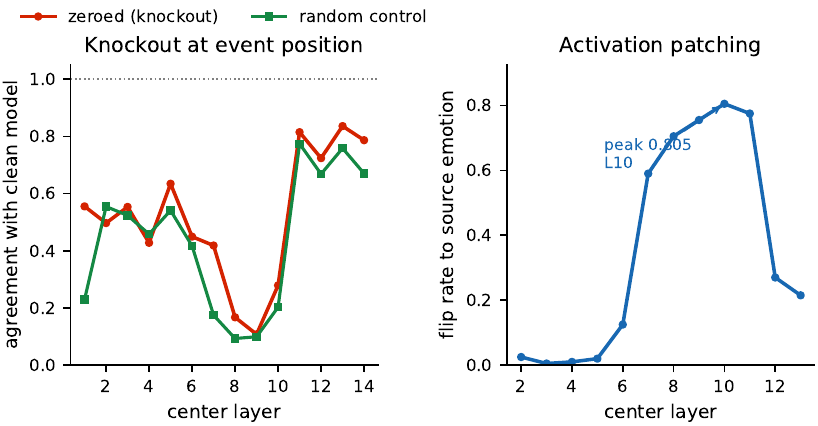}
  \caption{Residual-stream knockout and patching. Zeroing a 3-layer span at the event position (red) maximally
  disrupts the emotion decision; a random perturbation (green) is less damaging. Patching peaks at layer~10
  (flip rate $0.805$).}
  \label{fig:knockout}
\end{figure}

Emotion accuracy peaks at $0.943$ (linear) at layer~13 and the MLP at $0.947$ at layer~14; mean appraisal
$R^2$ reaches $0.30$ in the same band, led by \emph{pleasantness} ($0.70$) and \emph{unpleasantness}
($0.65$) (Figure~\ref{fig:layersweep}). The emotion geometry is dominated by valence: measured as the Pearson
correlation between $w_e^\top h$ and $w_a^\top h$, the strongest cells are emotions on pleasantness and
unpleasantness, and the agency contrast is weaker (anger loads $0.15$ on other- versus $-0.30$ on
self-responsibility; Figure~\ref{fig:emoapp}). Zeroing a 3-layer span at the event position collapses agreement
with the clean model to $0.11$ at layer~9, while a norm-matched random perturbation is less damaging; activation
patching peaks at $0.805$ at layer~10 (Figure~\ref{fig:knockout}).

\clearpage
\section{Behavioral Tests and In-Context Decodability}
\label{app:behavioral}
\begin{figure}[h]
  \centering
  \includegraphics[width=0.9\linewidth]{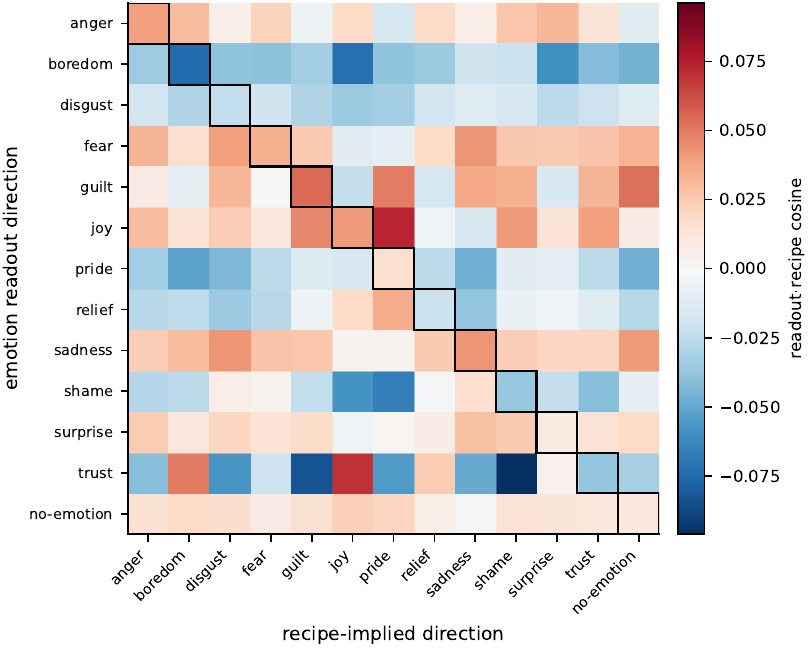}
  \caption{Emotion-to-recipe match matrix (Llama-3.2-1B). Rows are emotion readout directions, columns
  recipe-implied directions; a correct linear decomposition would put the maximum on the diagonal. Only a few
  emotions, led by anger, match their own recipe.}
  \label{fig:recipe}
\end{figure}

The behavioral tests use only generations and answer-token logits, so they can run on all four models. The 13 emotion-token logits are read for: (1) the emotion word, (2) its gold explication, (3) a same-affect decoy explication, and (4) a prime-scrambled control. Sufficiency ablates the explication one prime at a time (central vs peripheral by the gold recipe); the definitional floor scores the fraction of restatement tokens more frequent than the target. Table~\ref{tab:behav} has collected the numbers; Figure~\ref{fig:recipe} shows the full recipe-match matrix and Table~\ref{tab:decode} the in-context decodability behind the geometric non-reduction (only \textsc{feel} and \textsc{someone} clear $+0.10$).

\begin{table}[h]
  \centering\small
  \begin{tabular}{l r r r}
  \toprule
  Prime-direction source & rank-1 & rank-3 & $F_1$ \\
  \midrule
  random dictionary           & $0.08$ & $0.23$ & $0.00$ \\
  in-context (clean space)    & $0.08$ & $0.23$ & $0.35$ \\
  out-of-context (standalone) & $0.15$ & $0.39$ & $0.34$ \\
  \bottomrule
  \end{tabular}
  \caption{Linear emotion-to-recipe match for prime directions fit from the contrastive suite
  (Llama-3.2-1B). Chance is $0.08$ (rank-1) and $0.23$ (rank-3). The in-context source, the only one
  without a space mismatch, is at chance because the primes are consumed by the consolidation layer;
  linear composition is weak, in contrast to the behavioral reduction in Section~\ref{sec:reduce}.}
  \label{tab:decomp}
\end{table}

\begin{table}[t]
  \centering\small
  \begin{tabular}{l r r r}
  \toprule
  Prime & held-out & majority & lift \\
  \midrule
  \textsc{feel}    & $0.971$ & $0.745$ & $+0.226$ \\
  \textsc{someone} & $0.849$ & $0.739$ & $+0.109$ \\
  \textsc{not}     & $0.861$ & $0.778$ & $+0.084$ \\
  \textsc{i}       & $0.969$ & $0.939$ & $+0.030$ \\
  \textsc{because} & $0.894$ & $0.877$ & $+0.017$ \\
  \textsc{bad}     & $0.954$ & $0.970$ & $-0.015$ \\
  \bottomrule
  \end{tabular}
  \caption{In-context prime decodability at the consolidation layer, answer position (selected rows). Lift is
  held-out minus majority. Only \textsc{feel} and \textsc{someone} clear $+0.10$.}
  \label{tab:decode}
\end{table}

\end{document}